\documentclass{article}

\usepackage[final]{neurips_2025}

\usepackage[utf8]{inputenc} 
\usepackage[T1]{fontenc}    
\usepackage[colorlinks=true, linkcolor=blue, citecolor=blue, urlcolor=blue]{hyperref}
\usepackage{url}            
\usepackage{booktabs}       
\usepackage{amsfonts}       
\usepackage{nicefrac}       
\usepackage{microtype}      
\usepackage{xcolor}         

\usepackage{svg}
\usepackage{multirow}
\usepackage{multicol}
\usepackage{amsmath}
\usepackage{amssymb}
\usepackage{mathtools}
\usepackage{amsthm}

\usepackage{graphicx}
\usepackage{subfigure}
\usepackage{bbm}
\usepackage[capitalize,noabbrev]{cleveref}

\title{Analyzing Similarity Metrics for Data Selection for Language Model Pretraining}

\author{%
  Dylan Sam\thanks{Work done while interning at Google Research. Correspondence to \texttt{dylansam@andrew.cmu.edu}.} \\
  Carnegie Mellon University\\
  \And
  Ayan Chakrabarti \\
  Google Research \\
  \And
  Afshin Rostamizadeh \\
  Google Research \\
  \And
  Srikumar Ramalingam\\
  Google Research \\
  \And
  Gui Citovsky \\
  Google Research \\
  \And
  Sanjiv Kumar \\
  Google Research \\
}

\begin{document}

\maketitle

\begin{abstract}

Measuring similarity between training examples is critical for curating high-quality and diverse pretraining datasets for language models. 
However, similarity is typically computed with a generic off-the-shelf embedding model that has been trained for tasks such as retrieval. 
Whether these embedding-based similarity metrics are well-suited for pretraining data selection remains largely unexplored.
In this paper, we propose a new framework to assess the suitability of a similarity metric specifically for data curation in language model pretraining applications.
Our framework's first evaluation criterion captures how well distances reflect generalization in pretraining loss between different training examples.
Next, we use each embedding model to guide a standard diversity-based data curation algorithm and measure its utility by pretraining a language model on the selected data and evaluating downstream task performance.
Finally, we evaluate the capabilities of embeddings to distinguish between examples from different data sources.
With these evaluations, we demonstrate that standard off-the-shelf embedding models are not well-suited for the pretraining data curation setting, underperforming even remarkably simple embeddings that are extracted from models trained on the same pretraining corpus.
Our experiments are performed on the Pile, for pretraining a 1.7B parameter language model on 200B tokens.
We believe our analysis and evaluation framework serves as a foundation for the future design of embeddings that specifically reason about similarity in pretraining datasets.

\end{abstract}

\section{Introduction}

The recent success of language models~\citep{brown2020language, chowdhery2023palm} is in no small part due to pretraining on large and diverse text corpora scraped from a variety of sources~\citep{raffel2020exploring, gao2020pile, penedo2024fineweb}. 
Researchers have explored a variety of approaches to assemble effective pretraining sets, typically by selecting a high-quality and diverse subset from a larger corpus of examples scraped from multiple data sources.
These data curation approaches have shown promising results by improving example quality and reducing redundancy in pretraining sets. Many of these methods use notions of \emph{similarity} between examples. 
Similarity of an example to text from known high-quality sources (such as Wikipedia) has been used as a proxy for the quality of that example~\citep{gunasekar2023textbooks, penedo2024fineweb}.  
Meanwhile, methods focused on diversification~\citep{abbas2023semdedup, tirumala2023d4} make direct use of similarity metrics to identify and remove redundant examples. 

In this context, similarity between training examples has often been measured in terms of distances in an embedding space.
Many approaches have typically used generic embeddings for this purpose \citep{chang2023data, voautomatic}---off-the-shelf embedding models that have been trained for tasks such as semantic retrieval or mask-based reconstruction. 
However, whether these embeddings are optimal for reasoning about similarity between pretraining examples and whether there is any benefit to using more sophisticated and computationally expensive models remains an open question. 
Furthermore, evaluating such embeddings in pretraining applications is fundamentally challenging, as the sheer scale of pretraining data makes many experiments intractable.
Such challenges necessitate specialized evaluations --- new criteria for assessing the utility of embeddings in pretraining applications.

In this paper, we propose a novel evaluation framework to assess the suitability of an embedding model for curating pretraining data for language models. Our goal is to establish a \textbf{new standard} for evaluating similarity metrics in the context of language model pretraining.
We begin by asking: what should a similarity metric ideally capture to be useful in this setting? 
Unlike prior work that evaluates embeddings for downstream tasks like retrieval or classification, we argue that embeddings used for data curation must reflect training dynamics---such as generalization behavior or corpus redundancy---that are unique to the pretraining setting.

\begin{figure*}[t]
    \centering
    \includegraphics[width=0.95\textwidth]{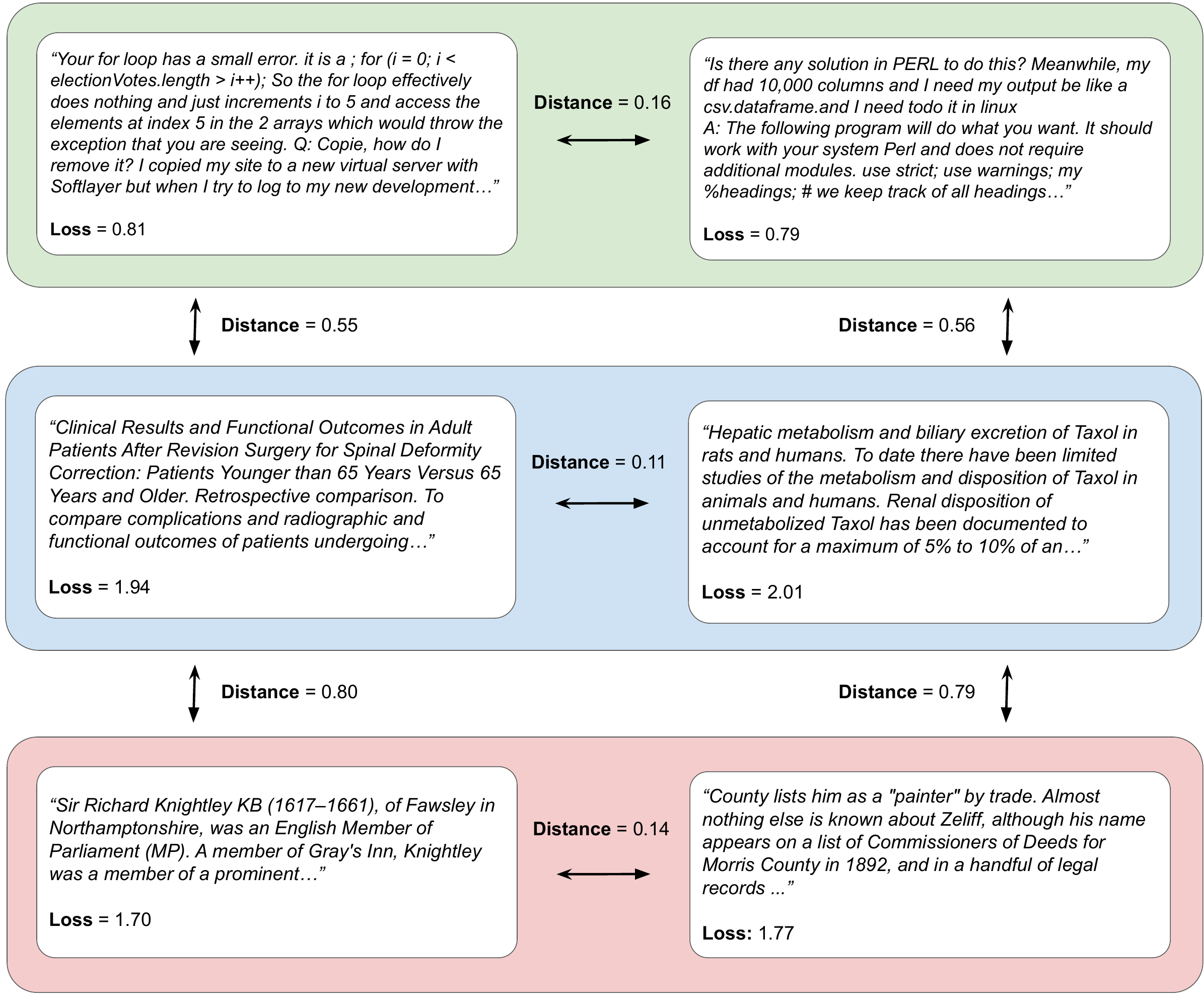}
    \caption{A visualization of the correlation between pretraining loss and embedding distance. Each row shows a pair of examples close in embedding space (from the same K-means cluster), with examples in different rows being far from each other (from different clusters). We find that close pairs of examples tend to have similar pretraining losses, while there is a greater variation in losses across clusters. Close example pairs are "thematically" similar but have different content. These results are from averaged embeddings from the final layer of a small decoder-only language model.}
    \label{fig:similarity_examples}
    \vspace{-4mm}
\end{figure*}

Our first evaluation criterion measures how well distances in the embedding space correlate with generalization behavior during pretraining---specifically, whether nearby examples in embedding space exhibit similar pretraining loss under the same model state (see Fig.~\ref{fig:similarity_examples}). This captures their utility of similarity metrics in loss-based data selection strategies \citep{jiang2023scaling} or for selecting examples similar to high-quality references \citep{penedo2024fineweb}.
Next, we evaluate the embedding model's utility in guiding a simple diversity-based curation method (similar to those employed in \citep{tirumala2023d4}) and assess its effectiveness by pretraining a model on the curated subset and measuring downstream task performance.
Finally, we also measure whether the embedding space can distinguish between examples from different data sources. While not directly used in data selection, this serves as a useful proxy for embedding quality under the assumption that these sources were curated by human expertise and reflect meaningful human-curated structure.

We conduct all of our experiments using the Pile~\citep{gao2020pile} as our data corpus, and in the context of pretraining a 1.7B parameter decoder-only language model with a UL2 objective~\citep{tay2022ul2} on 200B tokens. 
We evaluate a number of different off-the-shelf embedding models---from representations of generic language models (e.g., trained with mask-based reconstruction or mixtures of pretraining objectives) to embeddings specifically trained for retrieval or semantic equivalence. We compare these to two specialized approaches that are derived from smaller, proxy models of the downstream model being trained on the same pretraining set: (1) computing the average token embedding at the final hidden state, and (2) a significantly more computationally efficient and simple average of the token-embeddings from the input layer (i.e., requiring \textbf{no forward passes}). 

Despite their strong performance on standard semantic similarity and retrieval benchmarks, we find that off-the-shelf embedding models underperform in the context of pretraining data curation, suggesting that existing benchmarks may not reflect the inductive biases most relevant to language model pretraining. Surprisingly, even this simple method that averages input token embeddings---requiring no forward pass---matches or outperforms these more computationally intensive models. Adding a forward pass in this specialized model yields further improvements. 
These results strongly suggest that embedding models for reasoning about similarity between pretraining examples should be specialized to the data distribution at hand, and that properties that make embeddings suitable for retrieval and semantic matching may not transfer to the pretraining setting. 
More broadly, our work establishes a standardized framework to evaluate and highlight failures of current off-the-shelf models, facilitating the design of new models tailored for data curation in language model pretraining.

\section{Related Work}

\paragraph{Data Curation.} Many works have studied the problem of selecting high-quality and informative subsets from a larger corpus for efficient and effective language pretraining~\citep{Albalak2024ASO}. Indeed, works have shown that data curation for pretraining (both for language and vision-language models) improves neural scaling laws \citep{sorscher2022beyond}, and that curation techniques should be aware of the particular compute budget \citep{goyal2024scaling}.

One family of works approaches data curation with the goal of retaining only high quality examples. The simplest approaches in this family look at heuristics that can filter out noisy or extremely low-quality instances, such as documents containing less than 5 sentences or those with extremely short lines of text \citep{raffel2020exploring, xue2020mt5}. Beyond these heuristic approaches, a line of work focuses on extracting high-quality examples based on similarity to known high-quality sources, such as Wikipedia or textbooks~\citep{gunasekar2023textbooks, penedo2024fineweb}. 
Other works look at creating targeted sets---defining quality in terms of their relevance for particular downstream tasks \citep{xie2023data}, with some creating these by adaptively updating weights over domain mixtures \citep{xie2024doremi, jiang2024adaptive}. 
The other family of data curation approaches focus on pruning datasets to ensure a notion of \textit{diversity} and reducing redundancies in pretraining examples \citep{abbas2023semdedup, abbas2024effective, tirumala2023d4}. While some of these works~\citep{abbas2024effective, tirumala2023d4} compare different embedding models as part of brief ablations, these comparisons are in the narrow context of their specific diversification algorithms. 
Finally, it is worth mentioning that some works in data selection use the model being trained as part of the selection process, often through the use of influence functions \citep{garima2020estimating, xia2024less, engstrom2024dsdm}. 
However, these are typically only used in data-curation on small datasets or only during finetuning, and would be prohibitively expensive in pretraining.

\paragraph{Text Embedding Models.} Another related line of work is the task of learning embedding models for text. 
Many various approaches to learn text embedding models, with objectives including mask-based reconstruction \citep{devlin2018bert, liu2019roberta}, a combination of multiple different tasks \citep{cer2018universal}, and contrastive learning-based approaches \citep{gao2021simcse, neelakantan2022text, izacard2022unsupervised, lee2024gecko}.
More recent work has studied extracting embeddings from standard decoder-only language models \citep{behnamghader2024llm2vec}, even via prompting \citep{sam2025predicting}.

These embedding models have largely been studied in the context for classification or similarity measures \citep{gomaa2013survey, agirre2013sem, agirre2016semeval}, and recent focus has been on improving performance on aggregate on large-scale benchmarks \citep{muennighoff2022mteb} that are comprised of multiple tasks (e.g., retrieval, clustering, classification, ranking). 
Many models that achieve strong performance on these large-scale benchmarks have benefitted from scaling \citep{jiang2023scaling, chowdhery2023palm}, with the help of distilled information from autoregressive models \citep{lee2024gecko}. 
However, this has conflicting incentives with their utility in the pretraining setting, as large text embedding models are impractical and too computationally expensive to run inference over the full pretraining corpus. 
To the best of our knowledge, our work provides the first study of various text embedding models for pretraining data curation. 

\section{An Evaluation Framework for Embeddings in Pretraining Data Curation}

We now describe our new framework for analyzing text embedding models in terms of their suitability for reasoning about similarity among pretraining examples. 
As previously mentioned, similarity is used to (a) find examples that are similar to known ``desirable'' examples (e.g., examples of known high-quality or those representing downstream tasks of interest) and to (b) discover and remove redundancies in the pretraining corpus. 
Accordingly, we design standardized experiments that measure a text embedding model's performance towards these criteria.

\subsection{Evaluating Correlation with Pretraining Loss} \label{var_reduction}

We begin by evaluating whether low distances in a model's embedding space correlate with similar values of difficulty, which we measure by intermediate losses during language model pretraining. Examples of positive correlations are visualized in \Cref{fig:similarity_examples}.
To do so, we first use a balanced K-Means clustering algorithm in the embedding space to cluster all examples in the pretraining set for a target cluster size. Then, we look at the variance of loss values within each cluster. 
This within-cluster variance captures whether points with similar pretraining loss (or difficulty) are grouped together and are close in embedding space. Crucially, we note that the converse is not necessarily true; two very different examples can have the same loss because they happen to be equally difficult to learn. This asymmetry is an important reason we chose our cluster-based approach rather than alternatives such as pairwise correlation.
We note that we use balanced clustering (allowing only some variation in the sizes of different clusters) to ensure that average within-cluster variance is comparable across clusterings from different embedding models. 
We repeat this process for multiple target cluster sizes for all embeddings.

When reporting our results, we measure \textbf{variance reduction}, or the ratio of the overall variance across \textit{all examples} in the pretraining dataset to the variance in loss computed \textit{within clusters} in the given embedding space.
Formally, we define variance reduction as
\begin{align*}
    V(C) & = \frac{E_{x \sim D}[(\ell(x) - E[\ell(x)])^2]}{E_{C_i \sim \mathbbm{C}} \left[E_{x \sim C_i}[(\ell(x) - E_{x \sim C_i}[\ell(X)])^2 ] \right] },
\end{align*}
where $C = \{C_1, ..., C_m\}$ is a clustering or partition of all datapoints in the dataset $D$, and $\mathbbm{C}$ is a uniform distribution over each disjoint set in the partition $C$. 
Random clustering achieves a variance reduction of 1, and larger values of variance reduction imply that points with more similar pretraining loss are clustered together.

A high value of variance reduction implies that similarity in the embedding space correlates well with pretraining loss. While an isolated pair of examples having similar pretraining loss may be entirely unrelated in quality and content, the fact that an embedding space that \emph{consistently} brings together examples of similar loss values strongly suggests that these similar examples will behave similarly in terms of contributing to the quality of the language model (when used in pretraining).

Beyond finding examples of similar quality and utility, a high variance reduction score implies that the clustering above can serve as a particularly useful proxy for more dynamic and online data sampling strategies \citep{xie2024doremi}, reducing the number of required forward passes in strategies where datapoints are selected based on their current pretraining loss. 
This also has implications towards approaches that look to self-improve models through propagating labeled information to nearby examples \citep{wei2021theoretical, cai2021theory, pukdee2023label}, such as weak-to-strong generalization \citep{burns2023weak}. 

\subsection{Diversification-based Pretraining Data Curation} \label{sec:diverse_selection}

Prior work has demonstrated that data curation schemes that encourage diversification, based on similarity in embedding spaces, leads to improved models when trained on the curated subsets~\citep{abbas2023semdedup, tirumala2023d4}. 
While the diversification techniques tend to be sophisticated, our goal here is to evaluate the utility of a given embedding for diversification. 
Therefore, we use a simplified version of these approaches to (1) select a subset from a pretraining corpus, (2) pretrain a model on this curated subset, and then (3) report performance on a large set of downstream tasks.

We begin by clustering the larger pretraining corpus in the given embedding space; in contrast to \Cref{var_reduction}, we do not use balanced clustering, which can lead to clusters with a wide variation in ``diameters'' (distances among points in the same cluster). 
Instead, since the goal here is diversification, we find clusters such that distances between all pairs of points in the same cluster are within a specified threshold $\epsilon$.

Given clusters of pretraining data, prior work often uses complex pruning or sampling schemes from large clusters. For simplicity, we simply take the point in each cluster that is closest to the cluster centroid (i.e., the average embedding of all points in the cluster), which likely represents the most representative example for that cluster. 
We note that prior work \citep{tirumala2023d4} emphasizes the importance of a deduplication step in language model pretraining, and our curation strategy implicitly performs this process when selecting only a single point from each cluster.

Given that we select a single point from each cluster, this implies that we need the number of clusters to be equal to or greater than the number of desired pretraining examples. In practice, we sweep a large grid of $\epsilon$ values and select the largest threshold that still produces a sufficient number of clusters. 
Another important point of note is that producing such a large number of clusters is computationally expensive. 
To be able to scale to such a large number of clusters, we use the reciprocal agglomerative clustering (RAC) \citep{sumengen2021scaling} algorithm. 
This has the advantage of building a clustering from progressively larger values of $\epsilon$, which works well with our need for a grid of values of $\epsilon$. 

Finally, we pretrain a language model on the selected subset of examples, and measure few-shot performance of the pretrained model on a diverse variety of downstream tasks.

\subsection{Measuring Cluster Purity with Respect to Data Sources} \label{cluster_purity}

Large pretraining datasets such as the Pile \citep{gao2020pile} are comprised of various distinct yet complementary hand-curated data sources (e.g., high quality sources such as Wikipedia, code-based data, data from medical domains).
While there may be some similarities between datapoints from different domains, we believe that clusters should generally group points from the same domain together and separate instances from different domains.
In fact, since embedding models are generally not trained with explicit knowledge of source labels or any domain metadata, the ability to group together data from the same source and distinguish those from different sources shows an embedding model's alignment with meaningful human-curated structure (or at least that of the dataset curators).
This also serves as a useful proxy of embedding model capacity, as they should contain sufficient information to well-separate these different sources.

Letting $s \in \{0, \cdots, S \}$ represent the index of the source in some finite number of sources $S$, we compute the \textbf{cluster purity} of a set of clusters $C$ as
\begin{align*}
    P(C) & = \mathbbm{E}_{C_i \sim \mathbbm{C}}\left[ \frac{\max_s | C_i \cap D_s |}{|C_i|} \right],
\end{align*}
where $D_s$ represents the subset of datapoints that are from source $s$. Intuitively, this represents the proportion of the maximum frequency source to the total cluster size. A cluster purity of 1 represents a clustering that completely separates examples from different domains.

\section{Experiments}

We demonstrate the utility of our framework in evaluating off-the-shelf embedding models in measuring pretraining similarity and compare them with simple ways to produce specialized embeddings. 

\subsection{Experiment Details}

\paragraph{Embedding Models} For the off-the-shelf embedding models, we consider: (1) Universal Sentence Encoders (\textbf{USE}) \citep{cer2018universal}, which are general-purpose text encoders trained on a combination of objectives, (2) \textbf{Gecko} \citep{lee2024gecko}, a retrieval-focused text embedding model trained via synthetic data distilled from LLMs, and (3) \textbf{BERT} \citep{devlin2018bert} embedding models. 
These encapsulate a variety of different training objectives and reflect common embedding model choices in the field.

We also consider a few standard approaches to extract specialized embeddings that are extracted from a small version of the downstream language model we are training (e.g., a ~200M parameter language model). 
The first and most simple approach is to extract an embedding by simply averaging the token embeddings matrix over all tokens in the input sequence, which we refer to as \textbf{LM Token Embeds}. This is extremely efficient as it only requires looking up the token embedding matrix and \textbf{does not involve any forward passes}. We note that this is also equivalent to a learned unigram model (i.e., ignoring all positional information). 
The second approach is to extract an embedding from the forward pass of the language model by averaging the activations over all tokens in the input sequence at the final hidden layer, which we refer to as \textbf{LM Output Embeds}. 

Finally, in our diversity-based pretraining data curation experiments, we also add a comparison to a naive, random subset selection (\textbf{Baseline}). This evaluates how much benefit is observed from our standardized curation strategy with these various embedding models over standard baseline training.

\paragraph{Clustering Details} To make clustering with a large number of output clusters feasible and efficient at pretraining data scales, we perform dimensionality reduction using PCA on a large subset of the datapoints ($\sim$500,000 examples) and extract the top 64 dimensions upon which to project. 
The projected embeddings are then normalized to have a unit L2 norm.
We ablate on this choice for dimensionality reductions by comparing with random projections \citep{bingham2001random} as an alternative dimensionality reduction strategy; we find that random projections do not perform as well as PCA (\Cref{sec:ablations}). 
Other work \citep{tirumala2023d4} tackles this computational challenge by running clustering only with a much smaller subset of data, rather than performing dimensionality reduction and then clustering over all the data.

\begin{figure}[t]
    \centering
    \begin{minipage}[t]{0.49\textwidth}
        \centering
        \includegraphics[width=\textwidth]{figs/var_reduction_v3.pdf}
        \caption{Variance reduction as we vary average cluster size. Larger values are better. Results are computed over 50 million sampled clusters from the Pile, where pretraining losses are computed after 26k gradient steps. \textbf{Specialized embeddings yield higher variance reduction than off-the-shelf models for all cluster sizes.}}
        \label{fig:loss_clusters_kmeans}
    \end{minipage}
    \hfill
    \begin{minipage}[t]{0.49\textwidth}
        \centering
        \includegraphics[width=\textwidth]{figs/vary_ckpt_v3.pdf}
        \caption{Variance reduction as we increase the number of gradient steps in pretraining. Larger values are better. Results are computed over 50 million sampled clusters from the Pile with an average cluster size of 50. \textbf{Benefits in variance reduction remain consistent throughout pretraining}.}
        \label{fig:loss_clusters_ckpt}
    \end{minipage}
\end{figure}

\paragraph{Dataset and Pretraining Details}
For all of our experiments (e.g., pretraining data curation and predicting loss generalization), we use the Pile \citep{gao2020pile}. 
We pack together documents into a sequence of length 1280, with ``[eod]'' as delimiters between documents. For our pretraining experiments, we train a 1.7B parameter decoder-only language model with a UL2 objective \citep{tay2022ul2}. 
We curated pretraining subsets from the Pile through the process outlined in \Cref{sec:diverse_selection}. 
We use a selection budget of 200B tokens, or approximately 20\% of the Pile. 
This corresponds to roughly 170 million clusters for each embedding model, with an average cluster size of 5 examples. 

We pretrain with a learning rate of 0.001 with a linear decay and a batch size of 1024. 
For our tokenizer, we use sentencepiece with a vocabulary size of 256k tokens. 
For our pretraining evaluation, we consider the set of 23 downstream evaluation datasets from the work of \citet{brown2020language}. 
This involves a wide variety of 1-shot scoring tasks, as well as open-ended text generation tasks. 
More details about our evaluation sets are deferred to \Cref{appx:eval_details}. 
For the embeddings produced via LM Token Embeds and LM Output Embeds, we train a 200M parameter language model in the same fashion as above, where we also train on a total of 200B randomly selected tokens.

\begin{table*}[t]
    \centering
    \caption{Average downstream task performance of embedding models in the diversity-based curation schemes of a 200B token subset from the Pile, to pretrain a 1.7B parameter decoder-only language model. Bolding and italicizing denote the best and second-best performing methods on each task, respectively. Results are averaged over 3 pretraining runs, and average results are mean $\pm$ standard error. 
    \textbf{Specialized embeddings} (i.e., extracted from a small version of the downstream language model) \textbf{and Gecko perform the best for diversity-based pretraining data curation}.}
    \vspace{2mm}
    \resizebox{\textwidth}{!}{
\begin{tabular}{l|c | ccc | cc} \toprule
\multirow{2}{*}{\textbf{Task}}  & \multirow{2}{*}{\textbf{Baseline}}  & \multirow{2}{*}{\textbf{USE}}  & \multirow{2}{*}{\textbf{Gecko}} & \multirow{2}{*}{\textbf{BERT}} & \multirow{2}{*}{\shortstack{\textbf{LM Token}\\\textbf{Embeds}}} & \multirow{2}{*}{\shortstack{\textbf{LM Output}\\\textbf{Embeds}}} \\ 
& & & & & & \\ \midrule
ARC - Challenge    & 32.4            & 33.0         & \textbf{33.7}   & 32.7          & 32.5            & \textit{33.5}   \\
ARC - Easy         & 63.8            & \textit{65.1}         & 64.2          & 65.1          & 64.9            & \textbf{65.5}      \\
BoolQ              & 56.5            & 59.2         & 58.3          & 60.1          & \textbf{62.9}  & \textit{61.9}   \\
SuperGLUE - CB     & 42.4            & 43.5         & \textit{48.2} & 41.7          & 42.9            & \textbf{48.8}      \\
SuperGLUE - Copa   & 75.3            & 76.3         & \textit{77.3} & 74.0          & 74.7            & \textbf{78.0}      \\
HellaSwag          & 55.0            & 56.7         & \textit{57.1} & 56.5          & 56.8            & \textbf{57.5}      \\
Multi RC           & \textbf{57.7}   & 55.7         & 55.9          & \textit{56.3} & 56.0            & 55.2              \\
OpenBook QA        & 46.3            & 46.2         & 46.3          & \textit{46.6} & 45.9            & \textbf{46.7}      \\
PiQA               & 72.3            & 73.0         & 72.7          & \textit{73.6} & 72.8            & \textbf{73.6}      \\
Race H             & 38.0            & 37.9         & 38.5          & \textit{38.8} & \textbf{38.8}   & 38.7              \\
Race M             & 51.7            & 52.0         & \textbf{53.3}   & \textit{52.9} & 52.1            & 52.6              \\
ReCoRD             & \textbf{85.0}   & 84.4         & 84.6          & 84.7          & 84.5            & \textit{84.9}   \\
RTE                & \textit{54.5}& \textbf{54.9}& 51.5          & \textbf{54.9} & 51.6            & 52.8              \\
Story Cloze        & 73.9            & 74.1         & \textit{74.3} & 74.0          & 73.3            & \textbf{74.5}      \\
WiC                & \textbf{48.2}   & 47.9         & 47.5          & 47.5          & \textit{48.2} & 47.3              \\
Winograd           & 74.0            & 76.0         & \textit{77.1} & \textbf{77.3} & \textit{77.1} & 76.6              \\
WinoGrande         & 59.2            & 58.9         & 59.0          & \textit{59.5} & \textbf{60.0}   & 59.3              \\
WSC                & 74.1            & 73.9         & 73.7          & \textit{74.7} & 73.2            & \textbf{75.0}      \\
\midrule
Lambada            & 21.4            & 29.9         & \textbf{40.1}   & \textit{34.7} & 31.4            & 33.8              \\
Natural Questions  & 10.0            & 10.8         & 10.1          & 10.5          & \textit{11.1} & \textbf{11.4}      \\
SQuAD v2           & 51.1            & 54.3         & 51.7          & 51.8          & \textbf{58.6}   & \textit{55.4}   \\
TriviaQA Wiki      & 34.4            & \textit{35.0} & 33.0          & 35.0          & 34.3            & \textbf{36.1}      \\
Web Questions      & 17.1            & 17.0         & 17.9          & 18.0          & \textit{18.7} & \textbf{19.1}      \\
\midrule
\textbf{Average}   & 51.9 $\pm$ 0.1            & 52.9 $\pm$0.1         & \textit{53.3} $\pm$ \textit{0.3} & 53.1 $\pm$ 0.2          & 53.1 $\pm$ 0.3            & \textbf{53.8 $\pm$ 0.1}  \\
\bottomrule
\end{tabular}
    }
    \label{tab:pretraining_results}
\end{table*}

\subsection{Pretraining Loss Correlation Results}

To measure the ability of embedding models to reflect pretraining loss generalization and to cluster together datapoints with similar difficulty, we report their variance reduction as we vary (1) the number of gradient updates performed by the intermediate pretraining checkpoint and (2) as we vary the average cluster size in our K-Means clustering. 

We observe that specialized embeddings (i.e., LM Token Embeds and LM Output Embeds) from a small version of the downstream trained model, \textit{which is trained on the same pretraining data}, achieve a much larger reduction in loss variance when compared to off-the-shelf models. These specialized embeddings produce better clusters with datapoints with similar intermediate pretraining losses, across all cluster sizes (\Cref{fig:loss_clusters_kmeans}) and for all intermediate pretraining checkpoints (\Cref{fig:loss_clusters_ckpt}). 
This suggests that training models specifically on the data to perform data curation is preferable to using off-the-shelf embedding models for predicting pretraining loss generalization.
Furthermore, we also generally observe that newer embedding models (e.g., Gecko) do not improve upon older alternatives (e.g., BERT and USE), supporting that embedding model improvements in other settings, such as retrieval \textit{do not translate} to this application in the pretraining setting.

\paragraph{Visualizing Within-Cluster Examples} 

To better interpret these similarity metrics, we visualize pairs of datapoints from the clusters in \Cref{fig:similarity_examples} to understand what types of examples are placed closely together and which have similar pretraining losses. 
We randomly sample 3 clusters produced by K-Means clusters with the smallest average cluster size, with the LM Output Embeds method to extract embeddings. 
We randomly sample 2 points from these 3 clusters and observe that samples from the same cluster are thematically similar, although containing very different information (e.g., the last pair referring to two different government-related people, although from different times and different countries). 
We observe that pretraining losses are similar among within-cluster examples, and lower distances correspond to smaller differences in pretraining loss.

\subsection{Diversification-based Pretraining Data Curation Results}

We report the average downstream task performance of language models trained on pretraining data subsets produced by using various embedding models in the simple diversity-based data curation strategy in \Cref{tab:pretraining_results}. 
We first observe that the standardized diversity-based data curation technique applied to all embedding models (both off-the-shelf and those based on small downstream trained language models) outperforms the naive random subset selection baseline. 
Secondly, we observe that the LM Output Embeds achieves the best performance when averaged across all tasks. 

A notable result is that LM Token Embeds performs comparably to Gecko, while outperforming all other off-the-shelf embeddings. 
This is of immediate practical interest due to the simplicity of LM Token Embeds; they require no forward pass and, consequently, are extremely quick to compute in comparison to all other considered alternatives. 
Thus, LM Token Embeds serves as a viable alternative in settings that are compute-limited. 
This also suggests that even simple notions of similarity, even those that do not account for positional information, are sufficient for many pretraining data curation applications.
Finally, they support that models specialized to this task (e.g., trained on the same dataset) often outperform the current, general-purpose, off-the-shelf embedding models. While this does include training an additional model, it is a single-time investment that is amortized across all embedding use cases and is easily offset by the significant improvements in data curation quality.

\subsection{Cluster Purity Results}
\begin{figure}[t]
    \centering
    \includegraphics[width=0.49\columnwidth]{figs/purity_v3.pdf}
    \caption{Comparison of the purity with respect to data source of K-Means clustering produced by various embedding models on the Pile, when averaged over 50 million clusters from the Pile. \textbf{Specialized embedding models have higher cluster purity scores.}}
    \label{fig:purity}
\end{figure}

To better understand and interpret the similarity metrics defined by these embedding models, we measure the purity of clusters with respect to the underlying data source. 
We again remark that none of these embedding models have been trained with knowledge of the data source. 
Overall, we observe that most embedding models produce fairly pure clusters, where a majority of points come from the same underlying data source (\Cref{fig:purity}). 
This supports that these embeddings models are generally aligned with human judgment about differences between types of data. 

We observe that the embeddings extracted from the small version of the trained downstream language model achieve the highest cluster purity. 
We also note that Gecko embeddings achieve the third-highest cluster purity, whereas it performs poorly on the task of predicting loss generalization through variance reduction. 
This suggests that an improved ability to predict pretraining loss generalization cannot simply be explained via producing clusters that are more pure with respect to underlying data sources.
Future work could explore incorporating domain information or metadata into the embedding model's learning objective, as a form of weak supervision \citep{sam2023losses}.

\begin{figure*}[t]
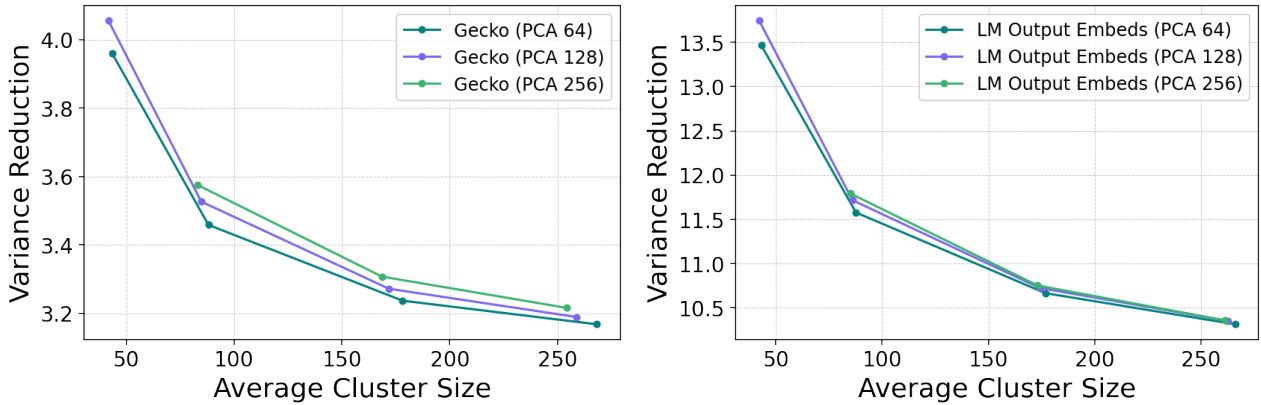

    \centering
    \includegraphics[width=0.49\textwidth]{figs/gecko_pca.pdf}
    \includegraphics[width=0.49\textwidth]{figs/lm_output_pca.pdf}
    \caption{Ablation on the number of components in PCA for Gecko and LM Output Embeds. Results are averaged over 50 million sampled clusters from the Pile. \textbf{Using more components in PCA better clusters points with similar pretraining loss.}}
    \label{fig:clustering_pca}
\end{figure*}

\begin{figure*}[t]
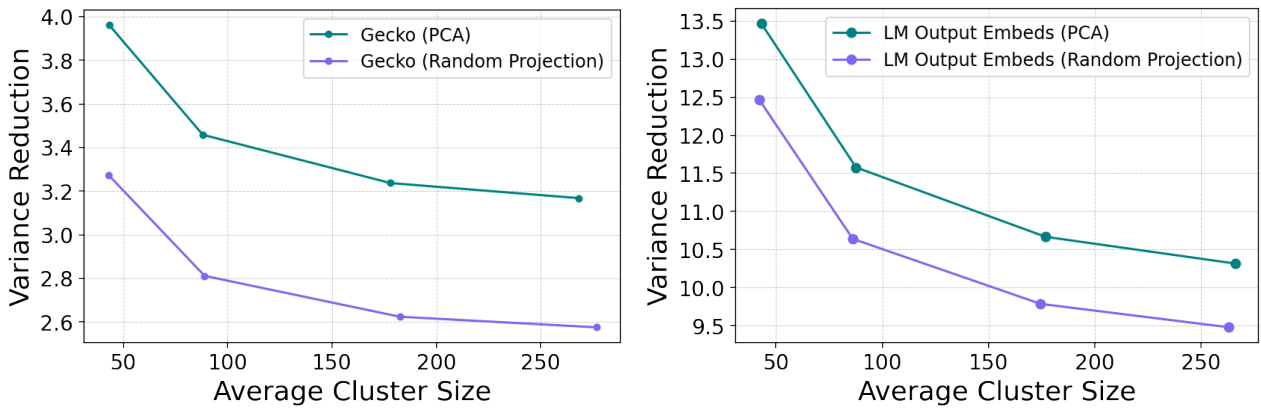

    \centering
    \includegraphics[width=0.49\textwidth]{figs/gecko_rp.pdf}
    \includegraphics[width=0.49\textwidth]{figs/lm_output_rp.pdf}
    \caption{Ablation comparing the use of PCA or Random projections for dimensionality reduction in Gecko and LM Output Embeds. Results are averaged over 50 million sampled clusters from the Pile. \textbf{Dimensionality reduction via PCA performs better than via Random Projections.}}
    \label{fig:random_projections}
\end{figure*}

\subsection{Ablations} \label{sec:ablations}

We present ablations on components of our evaluation framework --- specifically in how we have chosen to perform dimensionality reduction (i.e., technique and resulting size), which is used for the embedding models before running clustering. 
We present the ablations for Gecko and LM Output Embeds and defer results on others embeddings to Appendix \ref{appx:extra_results}.

\paragraph{Less Components in PCA Does Not Significantly Hurt Performance.} The dimensionality of the embeddings used in clustering often must be low for pretraining scales. Here, we run an ablation studying the role of dimensionality (i.e., the number of components in PCA onto which the embeddings are projected) in the variance reduction through clustering (\Cref{fig:clustering_pca}). 
We remark that scaling clustering to accommodate smaller average cluster sizes (i.e., more cluster centers) is intractable for embeddings that have high dimensionality. Thus, we can only report results on various embeddings with a large number (i.e., 256) of components in PCA with larger average cluster sizes.  
We observe the trend across all embeddings that a higher dimension and larger number of PCA components improves the embedding models' ability to cluster points by pretraining loss (see other embedding model results in Appendix \ref{appx:pca}).

\paragraph{Dimensionality Reduction with PCA Outperforms Random Projections.}
Another common technique to perform dimensionality reduction (especially with respect to maintaining pairwise distances and similarities) is to use random projections. 
We experiment with using random projections in our loss clustering experiments and observe that it is outperformed by PCA in terms of variance reduction across all cluster sizes (\Cref{fig:random_projections}). 

\section{Discussion}

We present a new evaluation framework for the understudied area of similarity metrics used in language model pretraining data curation. 
Using our framework, we show that off-the-shelf embeddings---despite their widespread use---often underperform even simple, specialized alternatives such as an average of token embeddings that requires no forward pass. 
This also suggests that practitioners should train their own embedding models, and it suffices to train them on a small fraction of the data ($\sim$20\%) and at a much smaller parameter scale.
While it may seem intuitive that embeddings tailored to the pretraining task would outperform generic ones, this has not been systematically demonstrated and is certainly not yet standard practice. 
Our framework both surfaces this gap and offers a practical tool for guiding the design of embedding models optimized for data curation, which can lead to significant improvements in data efficiency.

While we believe that our results are fairly general, future work could extend our findings to other pretraining corpora.
Beyond evaluation, our findings have broader implications for pretraining workflows—for example, in selecting task-specific finetuning data \citep{xia2024less}, or identifying synthetic examples that resemble real data \citep{meng2022generating, sam2024finetuning}, and even studying the impacts of scaling up embedding models in these tasks.
Overall, our results underscore the importance of using task-aligned similarity metrics in pretraining, and our framework provides a foundation for future research in building and optimizing the design of embedding models tailored to this critical step of the language modeling pipeline.

\paragraph{Limitations} One limitation of our work is that the empirical demonstration of our new evaluation framework is restricted to the Pile dataset. We believe this is fairly standard in the language model pretraining data curation literature to focus on a single pretraining corpus (e.g., as is done in \citet{tirumala2023d4}), due to the computational costs of both data curation algorithms and pretraining.

\bibliography{neurips_2025}

\begin{thebibliography}{64}
\providecommand{\natexlab}[1]{#1}
\providecommand{\url}[1]{\texttt{#1}}
\expandafter\ifx\csname urlstyle\endcsname\relax
  \providecommand{\doi}[1]{doi: #1}\else
  \providecommand{\doi}{doi: \begingroup \urlstyle{rm}\Url}\fi

\bibitem[Abbas et~al.(2023)Abbas, Tirumala, Simig, Ganguli, and Morcos]{abbas2023semdedup}
Amro Abbas, Kushal Tirumala, D{\'a}niel Simig, Surya Ganguli, and Ari~S Morcos.
\newblock Semdedup: Data-efficient learning at web-scale through semantic deduplication.
\newblock \emph{arXiv preprint arXiv:2303.09540}, 2023.

\bibitem[Abbas et~al.(2024)Abbas, Rusak, Tirumala, Brendel, Chaudhuri, and Morcos]{abbas2024effective}
Amro Kamal~Mohamed Abbas, Evgenia Rusak, Kushal Tirumala, Wieland Brendel, Kamalika Chaudhuri, and Ari~S. Morcos.
\newblock Effective pruning of web-scale datasets based on complexity of concept clusters.
\newblock In \emph{The Twelfth International Conference on Learning Representations}, 2024.
\newblock URL \url{https://openreview.net/forum?id=CtOA9aN8fr}.

\bibitem[Agirre et~al.(2013)Agirre, Cer, Diab, Gonzalez-Agirre, and Guo]{agirre2013sem}
Eneko Agirre, Daniel Cer, Mona Diab, Aitor Gonzalez-Agirre, and Weiwei Guo.
\newblock * sem 2013 shared task: Semantic textual similarity.
\newblock In \emph{Second joint conference on lexical and computational semantics (* SEM), volume 1: proceedings of the Main conference and the shared task: semantic textual similarity}, pages 32--43, 2013.

\bibitem[Agirre et~al.(2016)Agirre, Banea, Cer, Diab, Gonzalez~Agirre, Mihalcea, Rigau~Claramunt, and Wiebe]{agirre2016semeval}
Eneko Agirre, Carmen Banea, Daniel Cer, Mona Diab, Aitor Gonzalez~Agirre, Rada Mihalcea, German Rigau~Claramunt, and Janyce Wiebe.
\newblock Semeval-2016 task 1: Semantic textual similarity, monolingual and cross-lingual evaluation.
\newblock In \emph{SemEval-2016. 10th International Workshop on Semantic Evaluation; 2016 Jun 16-17; San Diego, CA. Stroudsburg (PA): ACL; 2016. p. 497-511.} ACL (Association for Computational Linguistics), 2016.

\bibitem[Albalak et~al.(2024)Albalak, Elazar, Xie, Longpre, Lambert, Wang, Muennighoff, Hou, Pan, Jeong, Raffel, Chang, Hashimoto, and Wang]{Albalak2024ASO}
Alon Albalak, Yanai Elazar, Sang~Michael Xie, Shayne Longpre, Nathan Lambert, Xinyi Wang, Niklas Muennighoff, Bairu Hou, Liangming Pan, Haewon Jeong, Colin Raffel, Shiyu Chang, Tatsunori Hashimoto, and William~Yang Wang.
\newblock A survey on data selection for language models.
\newblock \emph{ArXiv}, abs/2402.16827, 2024.

\bibitem[BehnamGhader et~al.(2024)BehnamGhader, Adlakha, Mosbach, Bahdanau, Chapados, and Reddy]{behnamghader2024llm2vec}
Parishad BehnamGhader, Vaibhav Adlakha, Marius Mosbach, Dzmitry Bahdanau, Nicolas Chapados, and Siva Reddy.
\newblock Llm2vec: Large language models are secretly powerful text encoders.
\newblock \emph{arXiv preprint arXiv:2404.05961}, 2024.

\bibitem[Berant et~al.(2013)Berant, Chou, Frostig, and Liang]{berant2013semantic}
Jonathan Berant, Andrew Chou, Roy Frostig, and Percy Liang.
\newblock Semantic parsing on freebase from question-answer pairs.
\newblock In \emph{Proceedings of the 2013 conference on empirical methods in natural language processing}, pages 1533--1544, 2013.

\bibitem[Bingham and Mannila(2001)]{bingham2001random}
Ella Bingham and Heikki Mannila.
\newblock Random projection in dimensionality reduction: applications to image and text data.
\newblock In \emph{Proceedings of the seventh ACM SIGKDD international conference on Knowledge discovery and data mining}, pages 245--250, 2001.

\bibitem[Bisk et~al.(2020)Bisk, Zellers, Gao, Choi, et~al.]{bisk2020piqa}
Yonatan Bisk, Rowan Zellers, Jianfeng Gao, Yejin Choi, et~al.
\newblock Piqa: Reasoning about physical commonsense in natural language.
\newblock In \emph{Proceedings of the AAAI conference on artificial intelligence}, 2020.

\bibitem[Brown et~al.(2020)Brown, Mann, Ryder, Subbiah, Kaplan, Dhariwal, Neelakantan, Shyam, Sastry, Askell, et~al.]{brown2020language}
Tom Brown, Benjamin Mann, Nick Ryder, Melanie Subbiah, Jared~D Kaplan, Prafulla Dhariwal, Arvind Neelakantan, Pranav Shyam, Girish Sastry, Amanda Askell, et~al.
\newblock Language models are few-shot learners.
\newblock \emph{Advances in neural information processing systems}, 33:\penalty0 1877--1901, 2020.

\bibitem[Burns et~al.(2023)Burns, Izmailov, Kirchner, Baker, Gao, Aschenbrenner, Chen, Ecoffet, Joglekar, Leike, et~al.]{burns2023weak}
Collin Burns, Pavel Izmailov, Jan~Hendrik Kirchner, Bowen Baker, Leo Gao, Leopold Aschenbrenner, Yining Chen, Adrien Ecoffet, Manas Joglekar, Jan Leike, et~al.
\newblock Weak-to-strong generalization: Eliciting strong capabilities with weak supervision.
\newblock \emph{arXiv preprint arXiv:2312.09390}, 2023.

\bibitem[Cai et~al.(2021)Cai, Gao, Lee, and Lei]{cai2021theory}
Tianle Cai, Ruiqi Gao, Jason Lee, and Qi~Lei.
\newblock A theory of label propagation for subpopulation shift.
\newblock In \emph{International Conference on Machine Learning}, pages 1170--1182. PMLR, 2021.

\bibitem[Cer(2018)]{cer2018universal}
D~Cer.
\newblock Universal sentence encoder.
\newblock \emph{arXiv preprint arXiv:1803.11175}, 2018.

\bibitem[Chang and Jia(2023)]{chang2023data}
Ting-Yun Chang and Robin Jia.
\newblock Data curation alone can stabilize in-context learning.
\newblock In \emph{Proceedings of the 61st Annual Meeting of the Association for Computational Linguistics (Volume 1: Long Papers)}, pages 8123--8144, 2023.

\bibitem[Chowdhery et~al.(2023)Chowdhery, Narang, Devlin, Bosma, Mishra, Roberts, Barham, Chung, Sutton, Gehrmann, et~al.]{chowdhery2023palm}
Aakanksha Chowdhery, Sharan Narang, Jacob Devlin, Maarten Bosma, Gaurav Mishra, Adam Roberts, Paul Barham, Hyung~Won Chung, Charles Sutton, Sebastian Gehrmann, et~al.
\newblock Palm: Scaling language modeling with pathways.
\newblock \emph{Journal of Machine Learning Research}, 24\penalty0 (240):\penalty0 1--113, 2023.

\bibitem[Clark et~al.(2019)Clark, Lee, Chang, Kwiatkowski, Collins, and Toutanova]{clark2019boolq}
Christopher Clark, Kenton Lee, Ming-Wei Chang, Tom Kwiatkowski, Michael Collins, and Kristina Toutanova.
\newblock Boolq: Exploring the surprising difficulty of natural yes/no questions.
\newblock In \emph{NAACL}, 2019.

\bibitem[Clark et~al.(2018)Clark, Cowhey, Etzioni, Khot, Sabharwal, Schoenick, and Tafjord]{allenai:arc}
Peter Clark, Isaac Cowhey, Oren Etzioni, Tushar Khot, Ashish Sabharwal, Carissa Schoenick, and Oyvind Tafjord.
\newblock Think you have solved question answering? try arc, the ai2 reasoning challenge.
\newblock \emph{arXiv:1803.05457v1}, 2018.

\bibitem[Dagan et~al.(2010)Dagan, Dolan, Magnini, and Roth]{dagan2010recognizing}
Ido Dagan, Bill Dolan, Bernardo Magnini, and Dan Roth.
\newblock Recognizing textual entailment: Rational, evaluation and approaches--erratum.
\newblock \emph{Natural Language Engineering}, 16\penalty0 (1):\penalty0 105--105, 2010.

\bibitem[Devlin(2018)]{devlin2018bert}
Jacob Devlin.
\newblock Bert: Pre-training of deep bidirectional transformers for language understanding.
\newblock \emph{arXiv preprint arXiv:1810.04805}, 2018.

\bibitem[Engstrom et~al.(2024)Engstrom, Feldmann, and Madry]{engstrom2024dsdm}
Logan Engstrom, Axel Feldmann, and Aleksander Madry.
\newblock Dsdm: Model-aware dataset selection with datamodels.
\newblock In \emph{Forty-first International Conference on Machine Learning}, 2024.
\newblock URL \url{https://openreview.net/forum?id=GC8HkKeH8s}.

\bibitem[Gao et~al.(2020)Gao, Biderman, Black, Golding, Hoppe, Foster, Phang, He, Thite, Nabeshima, et~al.]{gao2020pile}
Leo Gao, Stella Biderman, Sid Black, Laurence Golding, Travis Hoppe, Charles Foster, Jason Phang, Horace He, Anish Thite, Noa Nabeshima, et~al.
\newblock The pile: An 800gb dataset of diverse text for language modeling.
\newblock \emph{arXiv preprint arXiv:2101.00027}, 2020.

\bibitem[Gao et~al.(2021)Gao, Yao, and Chen]{gao2021simcse}
T~Gao, X~Yao, and Danqi Chen.
\newblock Simcse: Simple contrastive learning of sentence embeddings.
\newblock In \emph{EMNLP 2021-2021 Conference on Empirical Methods in Natural Language Processing, Proceedings}, 2021.

\bibitem[Garima et~al.(2020)Garima, Liu, Kale, and Sundararajan]{garima2020estimating}
Garima, Frederick Liu, Satyen Kale, and Mukund Sundararajan.
\newblock Estimating training data influence by tracing gradient descent.
\newblock In \emph{Proceedings of the 34th International Conference on Neural Information Processing Systems}, pages 19920--19930, 2020.

\bibitem[Gomaa et~al.(2013)Gomaa, Fahmy, et~al.]{gomaa2013survey}
Wael~H Gomaa, Aly~A Fahmy, et~al.
\newblock A survey of text similarity approaches.
\newblock \emph{international journal of Computer Applications}, 68\penalty0 (13):\penalty0 13--18, 2013.

\bibitem[Goyal et~al.(2024)Goyal, Maini, Lipton, Raghunathan, and Kolter]{goyal2024scaling}
Sachin Goyal, Pratyush Maini, Zachary~C Lipton, Aditi Raghunathan, and J~Zico Kolter.
\newblock Scaling laws for data filtering--data curation cannot be compute agnostic.
\newblock In \emph{Proceedings of the IEEE/CVF Conference on Computer Vision and Pattern Recognition}, pages 22702--22711, 2024.

\bibitem[Gunasekar et~al.(2023)Gunasekar, Zhang, Aneja, Mendes, Del~Giorno, Gopi, Javaheripi, Kauffmann, de~Rosa, Saarikivi, et~al.]{gunasekar2023textbooks}
Suriya Gunasekar, Yi~Zhang, Jyoti Aneja, Caio C{\'e}sar~Teodoro Mendes, Allie Del~Giorno, Sivakanth Gopi, Mojan Javaheripi, Piero Kauffmann, Gustavo de~Rosa, Olli Saarikivi, et~al.
\newblock Textbooks are all you need.
\newblock \emph{arXiv preprint arXiv:2306.11644}, 2023.

\bibitem[Izacard et~al.(2022)Izacard, Caron, Hosseini, Riedel, Bojanowski, Joulin, and Grave]{izacard2022unsupervised}
Gautier Izacard, Mathilde Caron, Lucas Hosseini, Sebastian Riedel, Piotr Bojanowski, Armand Joulin, and Edouard Grave.
\newblock Unsupervised dense information retrieval with contrastive learning.
\newblock \emph{Transactions on Machine Learning Research}, 2022.
\newblock ISSN 2835-8856.
\newblock URL \url{https://openreview.net/forum?id=jKN1pXi7b0}.

\bibitem[Jiang et~al.(2023)Jiang, Huang, Luan, Wang, and Zhuang]{jiang2023scaling}
Ting Jiang, Shaohan Huang, Zhongzhi Luan, Deqing Wang, and Fuzhen Zhuang.
\newblock Scaling sentence embeddings with large language models.
\newblock \emph{arXiv preprint arXiv:2307.16645}, 2023.

\bibitem[Jiang et~al.(2024)Jiang, Zhou, Feng, Malladi, and Kolter]{jiang2024adaptive}
Yiding Jiang, Allan Zhou, Zhili Feng, Sadhika Malladi, and J~Zico Kolter.
\newblock Adaptive data optimization: Dynamic sample selection with scaling laws.
\newblock \emph{arXiv preprint arXiv:2410.11820}, 2024.

\bibitem[Joshi et~al.(2017)Joshi, Choi, Weld, and Zettlemoyer]{joshi2017triviaqa}
Mandar Joshi, Eunsol Choi, Daniel~S Weld, and Luke Zettlemoyer.
\newblock Triviaqa: A large scale distantly supervised challenge dataset for reading comprehension.
\newblock In \emph{Proceedings of the 55th Annual Meeting of the Association for Computational Linguistics (Volume 1: Long Papers)}, pages 1601--1611, 2017.

\bibitem[Khashabi et~al.(2018)Khashabi, Chaturvedi, Roth, Upadhyay, and Roth]{MultiRC2018}
Daniel Khashabi, Snigdha Chaturvedi, Michael Roth, Shyam Upadhyay, and Dan Roth.
\newblock Looking beyond the surface:a challenge set for reading comprehension over multiple sentences.
\newblock In \emph{NAACL}, 2018.

\bibitem[Kwiatkowski et~al.(2019)Kwiatkowski, Palomaki, Redfield, Collins, Parikh, Alberti, Epstein, Polosukhin, Devlin, Lee, et~al.]{kwiatkowski2019natural}
Tom Kwiatkowski, Jennimaria Palomaki, Olivia Redfield, Michael Collins, Ankur Parikh, Chris Alberti, Danielle Epstein, Illia Polosukhin, Jacob Devlin, Kenton Lee, et~al.
\newblock Natural questions: a benchmark for question answering research.
\newblock \emph{Transactions of the Association for Computational Linguistics}, 7:\penalty0 453--466, 2019.

\bibitem[Lai et~al.(2017)Lai, Xie, Liu, Yang, and Hovy]{lai2017race}
Guokun Lai, Qizhe Xie, Hanxiao Liu, Yiming Yang, and Eduard Hovy.
\newblock Race: Large-scale reading comprehension dataset from examinations.
\newblock In \emph{Proceedings of the 2017 Conference on Empirical Methods in Natural Language Processing}, pages 785--794, 2017.

\bibitem[Lee et~al.(2024)Lee, Dai, Ren, Chen, Cer, Cole, Hui, Boratko, Kapadia, Ding, et~al.]{lee2024gecko}
Jinhyuk Lee, Zhuyun Dai, Xiaoqi Ren, Blair Chen, Daniel Cer, Jeremy~R Cole, Kai Hui, Michael Boratko, Rajvi Kapadia, Wen Ding, et~al.
\newblock Gecko: Versatile text embeddings distilled from large language models.
\newblock \emph{arXiv preprint arXiv:2403.20327}, 2024.

\bibitem[Levesque et~al.(2012)Levesque, Davis, and Morgenstern]{levesque2012winograd}
Hector Levesque, Ernest Davis, and Leora Morgenstern.
\newblock The winograd schema challenge.
\newblock In \emph{Thirteenth international conference on the principles of knowledge representation and reasoning}, 2012.

\bibitem[Liu et~al.(2019)Liu, Ott, Goyal, Du, Joshi, Chen, Levy, Lewis, Zettlemoyer, and Stoyanov]{liu2019roberta}
Yinhan Liu, Myle Ott, Naman Goyal, Jingfei Du, Mandar Joshi, Danqi Chen, Omer Levy, Mike Lewis, Luke Zettlemoyer, and Veselin Stoyanov.
\newblock Roberta: A robustly optimized bert pretraining approach.
\newblock \emph{arXiv e-prints}, pages arXiv--1907, 2019.

\bibitem[Meng et~al.(2022)Meng, Huang, Zhang, and Han]{meng2022generating}
Yu~Meng, Jiaxin Huang, Yu~Zhang, and Jiawei Han.
\newblock Generating training data with language models: Towards zero-shot language understanding.
\newblock \emph{Advances in Neural Information Processing Systems}, 35:\penalty0 462--477, 2022.

\bibitem[Mihaylov et~al.(2018)Mihaylov, Clark, Khot, and Sabharwal]{OpenBookQA2018}
Todor Mihaylov, Peter Clark, Tushar Khot, and Ashish Sabharwal.
\newblock Can a suit of armor conduct electricity? a new dataset for open book question answering.
\newblock In \emph{EMNLP}, 2018.

\bibitem[Mostafazadeh et~al.(2016)Mostafazadeh, Chambers, He, Parikh, Batra, Vanderwende, Kohli, and Allen]{mostafazadeh2016corpus}
Nasrin Mostafazadeh, Nathanael Chambers, Xiaodong He, Devi Parikh, Dhruv Batra, Lucy Vanderwende, Pushmeet Kohli, and James Allen.
\newblock A corpus and evaluation framework for deeper understanding of commonsense stories.
\newblock \emph{arXiv preprint arXiv:1604.01696}, 2016.

\bibitem[Muennighoff et~al.(2022)Muennighoff, Tazi, Magne, and Reimers]{muennighoff2022mteb}
Niklas Muennighoff, Nouamane Tazi, Lo{\"\i}c Magne, and Nils Reimers.
\newblock Mteb: Massive text embedding benchmark.
\newblock \emph{arXiv preprint arXiv:2210.07316}, 2022.

\bibitem[Neelakantan et~al.(2022)Neelakantan, Xu, Puri, Radford, Han, Tworek, Yuan, Tezak, Kim, Hallacy, et~al.]{neelakantan2022text}
Arvind Neelakantan, Tao Xu, Raul Puri, Alec Radford, Jesse~Michael Han, Jerry Tworek, Qiming Yuan, Nikolas Tezak, Jong~Wook Kim, Chris Hallacy, et~al.
\newblock Text and code embeddings by contrastive pre-training.
\newblock \emph{arXiv preprint arXiv:2201.10005}, 2022.

\bibitem[Paperno et~al.(2016)Paperno, Kruszewski, Lazaridou, Pham, Bernardi, Pezzelle, Baroni, Boleda, and Fern{\'a}ndez]{paperno2016lambada}
Denis Paperno, Germ{\'a}n Kruszewski, Angeliki Lazaridou, Ngoc-Quan Pham, Raffaella Bernardi, Sandro Pezzelle, Marco Baroni, Gemma Boleda, and Raquel Fern{\'a}ndez.
\newblock The lambada dataset: Word prediction requiring a broad discourse context.
\newblock In \emph{Proceedings of the 54th Annual Meeting of the Association for Computational Linguistics (Volume 1: Long Papers)}, pages 1525--1534, 2016.

\bibitem[Penedo et~al.(2024)Penedo, Kydl{\'\i}{\v{c}}ek, Lozhkov, Mitchell, Raffel, Von~Werra, Wolf, et~al.]{penedo2024fineweb}
Guilherme Penedo, Hynek Kydl{\'\i}{\v{c}}ek, Anton Lozhkov, Margaret Mitchell, Colin Raffel, Leandro Von~Werra, Thomas Wolf, et~al.
\newblock The fineweb datasets: Decanting the web for the finest text data at scale.
\newblock \emph{arXiv preprint arXiv:2406.17557}, 2024.

\bibitem[Pilehvar and Camacho-Collados(2018)]{pilehvar2018wic}
Mohammad~Taher Pilehvar and Jose Camacho-Collados.
\newblock Wic: the word-in-context dataset for evaluating context-sensitive meaning representations.
\newblock \emph{arXiv preprint arXiv:1808.09121}, 2018.

\bibitem[Pukdee et~al.(2023)Pukdee, Sam, Ravikumar, and Balcan]{pukdee2023label}
Rattana Pukdee, Dylan Sam, Pradeep~Kumar Ravikumar, and Nina Balcan.
\newblock Label propagation with weak supervision.
\newblock In \emph{The Eleventh International Conference on Learning Representations}, 2023.
\newblock URL \url{https://openreview.net/forum?id=aCuFa-RRqtI}.

\bibitem[Raffel et~al.(2020)Raffel, Shazeer, Roberts, Lee, Narang, Matena, Zhou, Li, and Liu]{raffel2020exploring}
Colin Raffel, Noam Shazeer, Adam Roberts, Katherine Lee, Sharan Narang, Michael Matena, Yanqi Zhou, Wei Li, and Peter~J Liu.
\newblock Exploring the limits of transfer learning with a unified text-to-text transformer.
\newblock \emph{Journal of machine learning research}, 21\penalty0 (140):\penalty0 1--67, 2020.

\bibitem[Rajpurkar et~al.(2018)Rajpurkar, Jia, and Liang]{rajpurkar2018know}
Pranav Rajpurkar, Robin Jia, and Percy Liang.
\newblock Know what you don’t know: Unanswerable questions for squad.
\newblock In \emph{Proceedings of the 56th Annual Meeting of the Association for Computational Linguistics (Volume 2: Short Papers)}, pages 784--789, 2018.

\bibitem[Sakaguchi et~al.(2021)Sakaguchi, Bras, Bhagavatula, and Choi]{sakaguchi2021winogrande}
Keisuke Sakaguchi, Ronan~Le Bras, Chandra Bhagavatula, and Yejin Choi.
\newblock Winogrande: An adversarial winograd schema challenge at scale.
\newblock \emph{Communications of the ACM}, 64\penalty0 (9):\penalty0 99--106, 2021.

\bibitem[Sam and Kolter(2023)]{sam2023losses}
Dylan Sam and J~Zico Kolter.
\newblock Losses over labels: Weakly supervised learning via direct loss construction.
\newblock In \emph{Proceedings of the AAAI conference on artificial intelligence}, volume~37, pages 9695--9703, 2023.

\bibitem[Sam et~al.(2024)Sam, Willmott, Semedo, and Kolter]{sam2024finetuning}
Dylan Sam, Devin Willmott, Joao~D Semedo, and J~Zico Kolter.
\newblock Finetuning clip to reason about pairwise differences.
\newblock \emph{arXiv preprint arXiv:2409.09721}, 2024.

\bibitem[Sam et~al.(2025)Sam, Finzi, and Kolter]{sam2025predicting}
Dylan Sam, Marc Finzi, and J~Zico Kolter.
\newblock Predicting the performance of black-box llms through self-queries.
\newblock \emph{arXiv preprint arXiv:2501.01558}, 2025.

\bibitem[Sorscher et~al.(2022)Sorscher, Geirhos, Shekhar, Ganguli, and Morcos]{sorscher2022beyond}
Ben Sorscher, Robert Geirhos, Shashank Shekhar, Surya Ganguli, and Ari Morcos.
\newblock Beyond neural scaling laws: beating power law scaling via data pruning.
\newblock \emph{Advances in Neural Information Processing Systems}, 35:\penalty0 19523--19536, 2022.

\bibitem[Sumengen et~al.(2021)Sumengen, Rajagopalan, Citovsky, Simcha, Bachem, Mitra, Blasiak, Liang, and Kumar]{sumengen2021scaling}
Baris Sumengen, Anand Rajagopalan, Gui Citovsky, David Simcha, Olivier Bachem, Pradipta Mitra, Sam Blasiak, Mason Liang, and Sanjiv Kumar.
\newblock Scaling hierarchical agglomerative clustering to billion-sized datasets.
\newblock \emph{arXiv preprint arXiv:2105.11653}, 2021.

\bibitem[Tay et~al.(2022)Tay, Dehghani, Tran, Garcia, Wei, Wang, Chung, Shakeri, Bahri, Schuster, et~al.]{tay2022ul2}
Yi~Tay, Mostafa Dehghani, Vinh~Q Tran, Xavier Garcia, Jason Wei, Xuezhi Wang, Hyung~Won Chung, Siamak Shakeri, Dara Bahri, Tal Schuster, et~al.
\newblock Ul2: Unifying language learning paradigms.
\newblock \emph{arXiv preprint arXiv:2205.05131}, 2022.

\bibitem[Tirumala et~al.(2023)Tirumala, Simig, Aghajanyan, and Morcos]{tirumala2023d4}
Kushal Tirumala, Daniel Simig, Armen Aghajanyan, and Ari Morcos.
\newblock D4: Improving llm pretraining via document de-duplication and diversification.
\newblock \emph{Advances in Neural Information Processing Systems}, 36:\penalty0 53983--53995, 2023.

\bibitem[Vo et~al.()Vo, Khalidov, Darcet, Moutakanni, Smetanin, Szafraniec, Touvron, Oquab, Joulin, Jegou, et~al.]{voautomatic}
Huy~V Vo, Vasil Khalidov, Timoth{\'e}e Darcet, Th{\'e}o Moutakanni, Nikita Smetanin, Marc Szafraniec, Hugo Touvron, Maxime Oquab, Armand Joulin, Herve Jegou, et~al.
\newblock Automatic data curation for self-supervised learning: A clustering-based approach.
\newblock \emph{Transactions on Machine Learning Research}.

\bibitem[Wang et~al.(2019)Wang, Pruksachatkun, Nangia, Singh, Michael, Hill, Levy, and Bowman]{wang2019superglue}
Alex Wang, Yada Pruksachatkun, Nikita Nangia, Amanpreet Singh, Julian Michael, Felix Hill, Omer Levy, and Samuel Bowman.
\newblock Superglue: A stickier benchmark for general-purpose language understanding systems.
\newblock \emph{Advances in neural information processing systems}, 32, 2019.

\bibitem[Wei et~al.(2021)Wei, Shen, Chen, and Ma]{wei2021theoretical}
Colin Wei, Kendrick Shen, Yining Chen, and Tengyu Ma.
\newblock Theoretical analysis of self-training with deep networks on unlabeled data.
\newblock In \emph{International Conference on Learning Representations}, 2021.
\newblock URL \url{https://openreview.net/forum?id=rC8sJ4i6kaH}.

\bibitem[Xia et~al.(2024)Xia, Malladi, Gururangan, Arora, and Chen]{xia2024less}
Mengzhou Xia, Sadhika Malladi, Suchin Gururangan, Sanjeev Arora, and Danqi Chen.
\newblock {LESS}: Selecting influential data for targeted instruction tuning.
\newblock In \emph{Forty-first International Conference on Machine Learning}, 2024.
\newblock URL \url{https://openreview.net/forum?id=PG5fV50maR}.

\bibitem[Xie et~al.(2023)Xie, Santurkar, Ma, and Liang]{xie2023data}
Sang~Michael Xie, Shibani Santurkar, Tengyu Ma, and Percy~S Liang.
\newblock Data selection for language models via importance resampling.
\newblock \emph{Advances in Neural Information Processing Systems}, 36:\penalty0 34201--34227, 2023.

\bibitem[Xie et~al.(2024)Xie, Pham, Dong, Du, Liu, Lu, Liang, Le, Ma, and Yu]{xie2024doremi}
Sang~Michael Xie, Hieu Pham, Xuanyi Dong, Nan Du, Hanxiao Liu, Yifeng Lu, Percy~S Liang, Quoc~V Le, Tengyu Ma, and Adams~Wei Yu.
\newblock Doremi: Optimizing data mixtures speeds up language model pretraining.
\newblock \emph{Advances in Neural Information Processing Systems}, 36, 2024.

\bibitem[Xue(2020)]{xue2020mt5}
L~Xue.
\newblock mt5: A massively multilingual pre-trained text-to-text transformer.
\newblock \emph{arXiv preprint arXiv:2010.11934}, 2020.

\bibitem[Zellers et~al.(2019)Zellers, Holtzman, Bisk, Farhadi, and Choi]{zellers2019hellaswag}
Rowan Zellers, Ari Holtzman, Yonatan Bisk, Ali Farhadi, and Yejin Choi.
\newblock Hellaswag: Can a machine really finish your sentence?
\newblock \emph{arXiv preprint arXiv:1905.07830}, 2019.

\bibitem[Zhang et~al.(2018)Zhang, Liu, Liu, Gao, Duh, and Van~Durme]{zhang2018record}
Sheng Zhang, Xiaodong Liu, Jingjing Liu, Jianfeng Gao, Kevin Duh, and Benjamin Van~Durme.
\newblock Record: Bridging the gap between human and machine commonsense reading comprehension.
\newblock \emph{arXiv preprint arXiv:1810.12885}, 2018.

\end{thebibliography}
\bibliographystyle{plainnat}

\newpage
\appendix
\onecolumn

\section{Additional Experimental Results} \label{appx:extra_results}

\subsection{Impact of Dimensionality Reduction on Variance Reduction} \label{appx:pca}

We now present the remaining results for other embedding models, when we ablate the number of components used in PCA for K-Means clustering, specifically when looking at the reduction in variance of pretraining loss of points within the same cluster. We observe that more components in PCA indeed help achieve higher variance reduction across all embedding models.

\begin{figure*}[h]
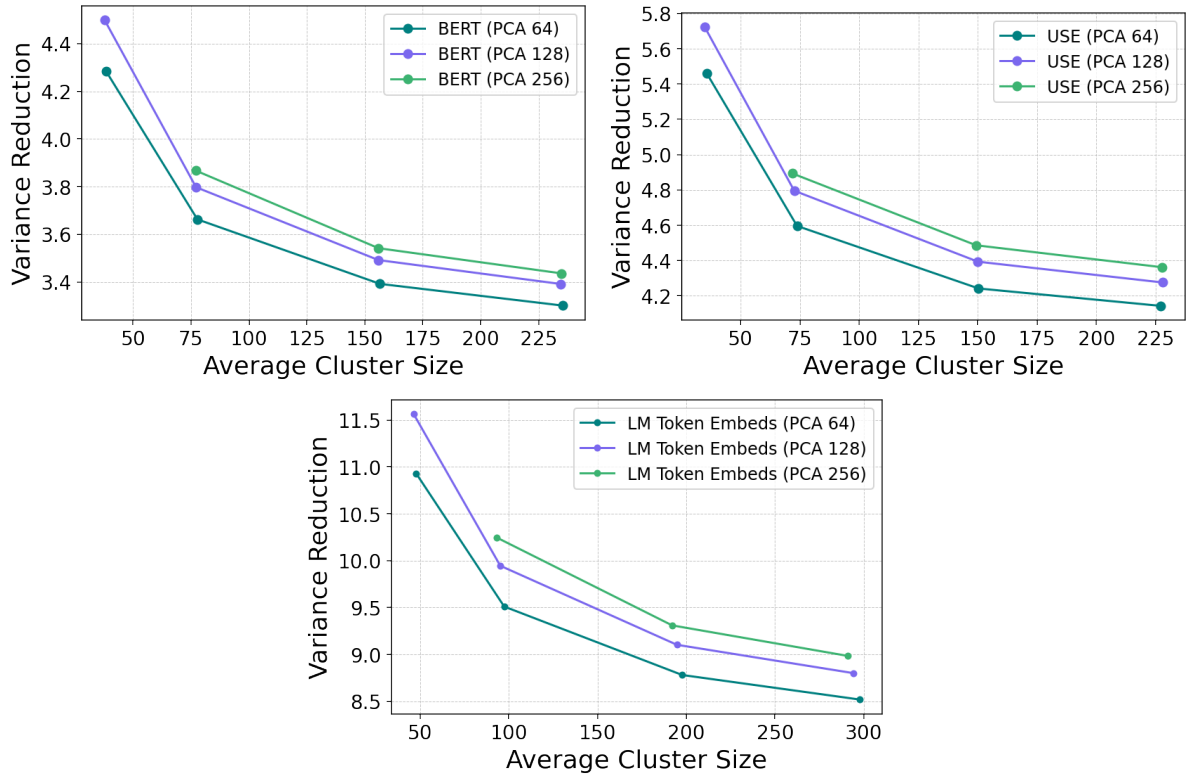

    \centering
    \includegraphics[width=0.49\textwidth]{figs/bert_pca.pdf}
    \includegraphics[width=0.49\textwidth]{figs/use_pca.pdf}
    \includegraphics[width=0.49\textwidth]{figs/lm_token_pca.pdf}
    \caption{Ablation on the number of components in PCA for embeddings from BERT, USE, and LM Token Embeds. Results are averaged over 50 million sampled clusters from the Pile.}
    \label{fig:clustering_pca_2}
\end{figure*}

\subsection{Comparison of Random Projections to PCA for Dimensionality Reduction} \label{appx:random_proj}

We now present the remaining results for other embedding models, when we use Random Projections for dimensionality reduction instead of using PCA. We consistently observe that embeddings paired with PCA outperform those using Random Projections.

\begin{figure*}[h]
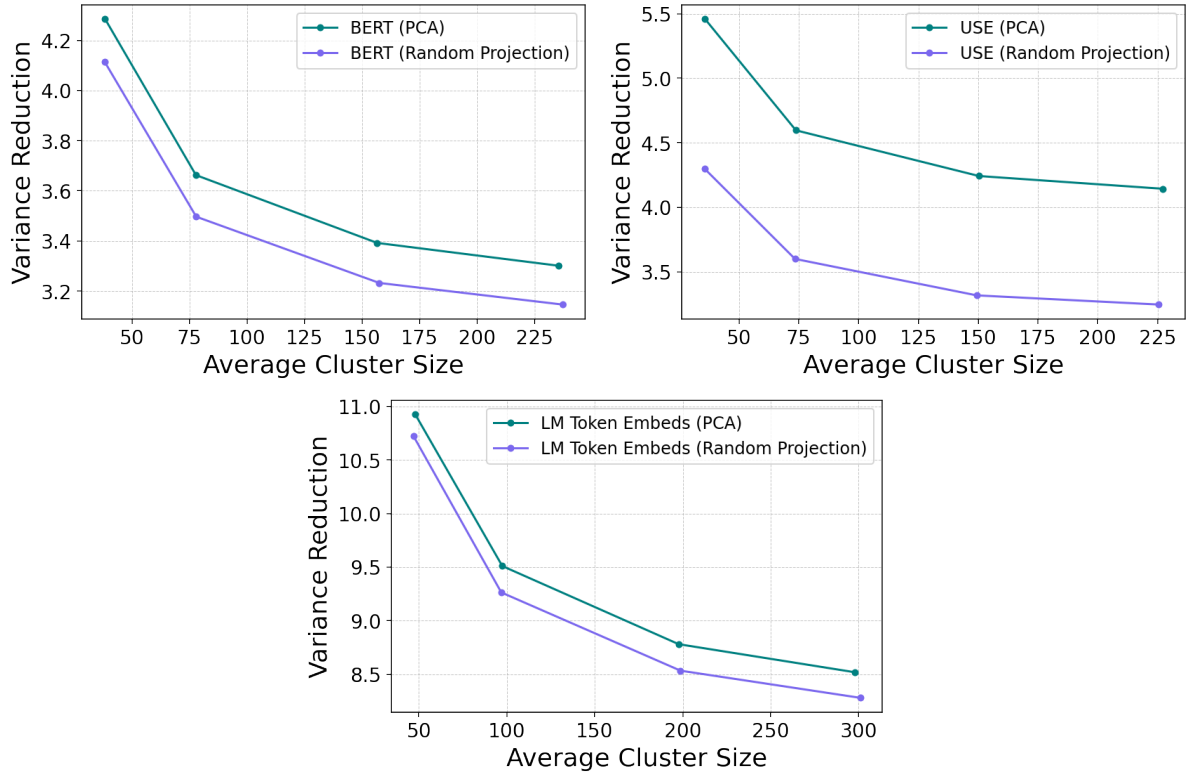

    \centering
    \includegraphics[width=0.49\textwidth]{figs/bert_rp.pdf}
    \includegraphics[width=0.49\textwidth]{figs/use_rp.pdf}
    \includegraphics[width=0.49\textwidth]{figs/lm_token_rp.pdf}
    \caption{Ablation where we compare using Random Projections for dimensionality reduction with embeddings from BERT, USE, and LM Token Embeds. Results are averaged over 50 million sampled clusters from the Pile.}
    \label{fig:random_proj_2}
\end{figure*}

\section{Additional Experimental Details}

\subsection{Additional Evaluation Details}\label{appx:eval_details}

We report results averaged over a large number of downstream tasks. These datasets largely follow two categories: scoring and decoding tasks. Scoring tasks primarily look at the output distribution of the model, while decoding tasks look at text generations from the language model. Scoring tasks are performed as 1-shot (i.e., giving one demonstration of format and answer), while decoding is performed zeroshot. For scoring tasks, we look at the standard top-1 accuracy. The list of scoring tasks is as follows:

\begin{itemize}
    \item ARC Challenge: Easy and Challenge \citep{allenai:arc} 
    \item BoolQ \citep{clark2019boolq}
    \item SuperGLUE - CB and Copa \citep{wang2019superglue}
    \item HellaSwag \citep{zellers2019hellaswag}
    \item MultiRC \citep{MultiRC2018}
    \item OpenBookQA \citep{OpenBookQA2018}
    \item PIQA \citep{bisk2020piqa}
    \item RACE-H and RACE-M \citep{lai2017race}
    \item ReCoRD \citep{zhang2018record}
    \item RTE \citep{dagan2010recognizing}
    \item Story Cloze \citep{mostafazadeh2016corpus}
    \item WIC \citep{pilehvar2018wic}
    \item Winograd \citep{levesque2012winograd}
    \item WinoGrande \citep{sakaguchi2021winogrande}
    \item WSC \citep{levesque2012winograd}
\end{itemize}

For decoding or text generation tasks, we evaluate the language model outputs with its F1 score. Decoder tasks are also evaluated as 1-shot. The list of decoding tasks is as follows:
\begin{itemize}
    \item Lambada \citep{paperno2016lambada}
    \item Natural Questions \citep{kwiatkowski2019natural}
    \item Squad v2 \citep{rajpurkar2018know}
    \item Trivia QA Wiki \citep{joshi2017triviaqa}
    \item Web Questions \citep{berant2013semantic}
\end{itemize}

\subsection{Additional Embedding Model Details}

The embedding models of BERT, Gecko and USE are trained with sequence lengths of 512, which we apply on the first 512 tokens of data from the Pile. 
For Gecko, we use the 110 million parameter model version, while for USE we use the 109 million parameter version. 
For BERT, we also use a 109 million parameter model. 
For the small language model that we train with the UL2 objective, we use one with approximately 200 million parameters. 

Both the off-the-shelf BERT and USE embedding models have a dimensionality of 512. The Gecko embedding model has a dimensionality of 768.
The small language model has a token embedding dimension of 512 and an hidden activation dimension of 512, which means that both LM Token Embeds and LM Output Embeds have 512 dimensions.

\subsection{Additional Hyperparameter Details}\label{appx:hyperparam_details}

\paragraph{Clustering}

For performing RAC clustering for our pretraining experiments, we use a value of $\epsilon$ as the particular diameter of clusters:
\begin{itemize}
    \item USE: $\epsilon = 0.2$, which defines roughly 225 million clusters
    \item Gecko: $\epsilon = 0.2$, which defines roughly 220 million clusters
    \item BERT: $\epsilon = 0.001$, which defines roughly 175 million clusters
    \item LM Token Embeds: $\epsilon = 0.001$, which defines roughly 170 million clusters
    \item LM Output Embeds: $\epsilon = 0.03$, which defines roughly 180 million clusters
\end{itemize}

For our K-Means clustering, we perform clustering at 4 different levels of granularity in our variance reduction and cluster purity results. We create four sets of clusterings with an average cluster size of 25, 50, 100, 150, with a minimum cluster size of $\frac{1}{5}$ times the average cluster size, and a maximum cluster size of $5$ times the average cluster size. 
For both RAC and K-Means, we use the squared L2 (Euclidean) distance.

\subsection{Dimensionality Reduction Details}

For running our dimensionality reduction via PCA, we compute the means and components on which to project on a subset of the data ($\sim$500,000 points). 
We first standardize the embeddings to have a mean of 0 and variance of 1 before running PCA. As previously mentioned, after projecting onto the desired number of principal components, we perform L2 normalization.

For our random projections, we use a sparse random projections onto values of $-\frac{\sqrt{n}}{\sqrt{64}}, 0, \frac{\sqrt{n}}{\sqrt{64}}$ with probabilities $\frac{1}{4}, \frac{1}{2}, \frac{1}{4}$ respectively (i.e., the default parameters via scikit-learn). We also L2 normalize the result of random projections.

\subsection{Compute Details}

Pretraining experiments for our 1.7B parameter language models are run on 512 v5 TPUs, where each pretraining run takes approximately 3 days. Training our proxy 200M parameter model took less than 1 day on 64 v5 TPUs. Hierarchical clustering for pretraining requires approximately 1-2 days to run over the full pretraining corpus.

\subsection{Asset Licenses}

The existing assets that we use have the following licenses:
\begin{itemize}
    \item RAC Clustering: MIT license
    \item The Pile: CC BY 4.0
    \item Evaluation Datasets: MIT License
\end{itemize}

\newpage
\section*{NeurIPS Paper Checklist}

\begin{enumerate}

\item {\bf Claims}
    \item[] Question: Do the main claims made in the abstract and introduction accurately reflect the paper's contributions and scope?
    \item[] Answer: \answerYes{} 
    \item[] Justification: All claims are supported by experimental and theoretical results.
    \item[] Guidelines:
    \begin{itemize}
        \item The answer NA means that the abstract and introduction do not include the claims made in the paper.
        \item The abstract and/or introduction should clearly state the claims made, including the contributions made in the paper and important assumptions and limitations. A No or NA answer to this question will not be perceived well by the reviewers. 
        \item The claims made should match theoretical and experimental results, and reflect how much the results can be expected to generalize to other settings. 
        \item It is fine to include aspirational goals as motivation as long as it is clear that these goals are not attained by the paper. 
    \end{itemize}

\item {\bf Limitations}
    \item[] Question: Does the paper discuss the limitations of the work performed by the authors?
    \item[] Answer: \answerYes{} 
    \item[] Justification: Limitations are discussed in the Discussion section.
    \item[] Guidelines:
    \begin{itemize}
        \item The answer NA means that the paper has no limitation while the answer No means that the paper has limitations, but those are not discussed in the paper. 
        \item The authors are encouraged to create a separate "Limitations" section in their paper.
        \item The paper should point out any strong assumptions and how robust the results are to violations of these assumptions (e.g., independence assumptions, noiseless settings, model well-specification, asymptotic approximations only holding locally). The authors should reflect on how these assumptions might be violated in practice and what the implications would be.
        \item The authors should reflect on the scope of the claims made, e.g., if the approach was only tested on a few datasets or with a few runs. In general, empirical results often depend on implicit assumptions, which should be articulated.
        \item The authors should reflect on the factors that influence the performance of the approach. For example, a facial recognition algorithm may perform poorly when image resolution is low or images are taken in low lighting. Or a speech-to-text system might not be used reliably to provide closed captions for online lectures because it fails to handle technical jargon.
        \item The authors should discuss the computational efficiency of the proposed algorithms and how they scale with dataset size.
        \item If applicable, the authors should discuss possible limitations of their approach to address problems of privacy and fairness.
        \item While the authors might fear that complete honesty about limitations might be used by reviewers as grounds for rejection, a worse outcome might be that reviewers discover limitations that aren't acknowledged in the paper. The authors should use their best judgment and recognize that individual actions in favor of transparency play an important role in developing norms that preserve the integrity of the community. Reviewers will be specifically instructed to not penalize honesty concerning limitations.
    \end{itemize}

\item {\bf Theory assumptions and proofs}
    \item[] Question: For each theoretical result, does the paper provide the full set of assumptions and a complete (and correct) proof?
    \item[] Answer: \answerNA{} 
    \item[] Justification: No theoretical results are included.
    \item[] Guidelines:
    \begin{itemize}
        \item The answer NA means that the paper does not include theoretical results. 
        \item All the theorems, formulas, and proofs in the paper should be numbered and cross-referenced.
        \item All assumptions should be clearly stated or referenced in the statement of any theorems.
        \item The proofs can either appear in the main paper or the supplemental material, but if they appear in the supplemental material, the authors are encouraged to provide a short proof sketch to provide intuition. 
        \item Inversely, any informal proof provided in the core of the paper should be complemented by formal proofs provided in appendix or supplemental material.
        \item Theorems and Lemmas that the proof relies upon should be properly referenced. 
    \end{itemize}

    \item {\bf Experimental result reproducibility}
    \item[] Question: Does the paper fully disclose all the information needed to reproduce the main experimental results of the paper to the extent that it affects the main claims and/or conclusions of the paper (regardless of whether the code and data are provided or not)?
    \item[] Answer: \answerYes{} 
    \item[] Justification: All experimental details are provided.
    \item[] Guidelines:
    \begin{itemize}
        \item The answer NA means that the paper does not include experiments.
        \item If the paper includes experiments, a No answer to this question will not be perceived well by the reviewers: Making the paper reproducible is important, regardless of whether the code and data are provided or not.
        \item If the contribution is a dataset and/or model, the authors should describe the steps taken to make their results reproducible or verifiable. 
        \item Depending on the contribution, reproducibility can be accomplished in various ways. For example, if the contribution is a novel architecture, describing the architecture fully might suffice, or if the contribution is a specific model and empirical evaluation, it may be necessary to either make it possible for others to replicate the model with the same dataset, or provide access to the model. In general. releasing code and data is often one good way to accomplish this, but reproducibility can also be provided via detailed instructions for how to replicate the results, access to a hosted model (e.g., in the case of a large language model), releasing of a model checkpoint, or other means that are appropriate to the research performed.
        \item While NeurIPS does not require releasing code, the conference does require all submissions to provide some reasonable avenue for reproducibility, which may depend on the nature of the contribution. For example
        \begin{enumerate}
            \item If the contribution is primarily a new algorithm, the paper should make it clear how to reproduce that algorithm.
            \item If the contribution is primarily a new model architecture, the paper should describe the architecture clearly and fully.
            \item If the contribution is a new model (e.g., a large language model), then there should either be a way to access this model for reproducing the results or a way to reproduce the model (e.g., with an open-source dataset or instructions for how to construct the dataset).
            \item We recognize that reproducibility may be tricky in some cases, in which case authors are welcome to describe the particular way they provide for reproducibility. In the case of closed-source models, it may be that access to the model is limited in some way (e.g., to registered users), but it should be possible for other researchers to have some path to reproducing or verifying the results.
        \end{enumerate}
    \end{itemize}

\item {\bf Open access to data and code}
    \item[] Question: Does the paper provide open access to the data and code, with sufficient instructions to faithfully reproduce the main experimental results, as described in supplemental material?
    \item[] Answer: \answerNo{} 
    \item[] Justification: No code is provided, but all experimental details and evaluation metrics are clearly defined.
    \item[] Guidelines:
    \begin{itemize}
        \item The answer NA means that paper does not include experiments requiring code.
        \item Please see the NeurIPS code and data submission guidelines (\url{https://nips.cc/public/guides/CodeSubmissionPolicy}) for more details.
        \item While we encourage the release of code and data, we understand that this might not be possible, so “No” is an acceptable answer. Papers cannot be rejected simply for not including code, unless this is central to the contribution (e.g., for a new open-source benchmark).
        \item The instructions should contain the exact command and environment needed to run to reproduce the results. See the NeurIPS code and data submission guidelines (\url{https://nips.cc/public/guides/CodeSubmissionPolicy}) for more details.
        \item The authors should provide instructions on data access and preparation, including how to access the raw data, preprocessed data, intermediate data, and generated data, etc.
        \item The authors should provide scripts to reproduce all experimental results for the new proposed method and baselines. If only a subset of experiments are reproducible, they should state which ones are omitted from the script and why.
        \item At submission time, to preserve anonymity, the authors should release anonymized versions (if applicable).
        \item Providing as much information as possible in supplemental material (appended to the paper) is recommended, but including URLs to data and code is permitted.
    \end{itemize}

\item {\bf Experimental setting/details}
    \item[] Question: Does the paper specify all the training and test details (e.g., data splits, hyperparameters, how they were chosen, type of optimizer, etc.) necessary to understand the results?
    \item[] Answer: \answerYes{} 
    \item[] Justification: All experimental details are provided in the code in the Appendix.
    \item[] Guidelines:
    \begin{itemize}
        \item The answer NA means that the paper does not include experiments.
        \item The experimental setting should be presented in the core of the paper to a level of detail that is necessary to appreciate the results and make sense of them.
        \item The full details can be provided either with the code, in appendix, or as supplemental material.
    \end{itemize}

\item {\bf Experiment statistical significance}
    \item[] Question: Does the paper report error bars suitably and correctly defined or other appropriate information about the statistical significance of the experiments?
    \item[] Answer: \answerYes{} 
    \item[] Justification: Error bars are provided when applicable.
    \item[] Guidelines:
    \begin{itemize}
        \item The answer NA means that the paper does not include experiments.
        \item The authors should answer "Yes" if the results are accompanied by error bars, confidence intervals, or statistical significance tests, at least for the experiments that support the main claims of the paper.
        \item The factors of variability that the error bars are capturing should be clearly stated (for example, train/test split, initialization, random drawing of some parameter, or overall run with given experimental conditions).
        \item The method for calculating the error bars should be explained (closed form formula, call to a library function, bootstrap, etc.)
        \item The assumptions made should be given (e.g., Normally distributed errors).
        \item It should be clear whether the error bar is the standard deviation or the standard error of the mean.
        \item It is OK to report 1-sigma error bars, but one should state it. The authors should preferably report a 2-sigma error bar than state that they have a 96\% CI, if the hypothesis of Normality of errors is not verified.
        \item For asymmetric distributions, the authors should be careful not to show in tables or figures symmetric error bars that would yield results that are out of range (e.g. negative error rates).
        \item If error bars are reported in tables or plots, The authors should explain in the text how they were calculated and reference the corresponding figures or tables in the text.
    \end{itemize}

\item {\bf Experiments compute resources}
    \item[] Question: For each experiment, does the paper provide sufficient information on the computer resources (type of compute workers, memory, time of execution) needed to reproduce the experiments?
    \item[] Answer: \answerYes{} 
    \item[] Justification: Compute details are provided in the Appendix.
    \item[] Guidelines:
    \begin{itemize}
        \item The answer NA means that the paper does not include experiments.
        \item The paper should indicate the type of compute workers CPU or GPU, internal cluster, or cloud provider, including relevant memory and storage.
        \item The paper should provide the amount of compute required for each of the individual experimental runs as well as estimate the total compute. 
        \item The paper should disclose whether the full research project required more compute than the experiments reported in the paper (e.g., preliminary or failed experiments that didn't make it into the paper). 
    \end{itemize}
    
\item {\bf Code of ethics}
    \item[] Question: Does the research conducted in the paper conform, in every respect, with the NeurIPS Code of Ethics \url{https://neurips.cc/public/EthicsGuidelines}?
    \item[] Answer: \answerYes{} 
    \item[] Justification: The submission conforms with the code of ethics.
    \item[] Guidelines:
    \begin{itemize}
        \item The answer NA means that the authors have not reviewed the NeurIPS Code of Ethics.
        \item If the authors answer No, they should explain the special circumstances that require a deviation from the Code of Ethics.
        \item The authors should make sure to preserve anonymity (e.g., if there is a special consideration due to laws or regulations in their jurisdiction).
    \end{itemize}

\item {\bf Broader impacts}
    \item[] Question: Does the paper discuss both potential positive societal impacts and negative societal impacts of the work performed?
    \item[] Answer: \answerYes{} 
    \item[] Justification: Impacts of the paper are discussed in the discussion section.
    \item[] Guidelines:
    \begin{itemize}
        \item The answer NA means that there is no societal impact of the work performed.
        \item If the authors answer NA or No, they should explain why their work has no societal impact or why the paper does not address societal impact.
        \item Examples of negative societal impacts include potential malicious or unintended uses (e.g., disinformation, generating fake profiles, surveillance), fairness considerations (e.g., deployment of technologies that could make decisions that unfairly impact specific groups), privacy considerations, and security considerations.
        \item The conference expects that many papers will be foundational research and not tied to particular applications, let alone deployments. However, if there is a direct path to any negative applications, the authors should point it out. For example, it is legitimate to point out that an improvement in the quality of generative models could be used to generate deepfakes for disinformation. On the other hand, it is not needed to point out that a generic algorithm for optimizing neural networks could enable people to train models that generate Deepfakes faster.
        \item The authors should consider possible harms that could arise when the technology is being used as intended and functioning correctly, harms that could arise when the technology is being used as intended but gives incorrect results, and harms following from (intentional or unintentional) misuse of the technology.
        \item If there are negative societal impacts, the authors could also discuss possible mitigation strategies (e.g., gated release of models, providing defenses in addition to attacks, mechanisms for monitoring misuse, mechanisms to monitor how a system learns from feedback over time, improving the efficiency and accessibility of ML).
    \end{itemize}
    
\item {\bf Safeguards}
    \item[] Question: Does the paper describe safeguards that have been put in place for responsible release of data or models that have a high risk for misuse (e.g., pretrained language models, image generators, or scraped datasets)?
    \item[] Answer: \answerNA{} 
    \item[] Justification: No such risks are posed.
    \item[] Guidelines:
    \begin{itemize}
        \item The answer NA means that the paper poses no such risks.
        \item Released models that have a high risk for misuse or dual-use should be released with necessary safeguards to allow for controlled use of the model, for example by requiring that users adhere to usage guidelines or restrictions to access the model or implementing safety filters. 
        \item Datasets that have been scraped from the Internet could pose safety risks. The authors should describe how they avoided releasing unsafe images.
        \item We recognize that providing effective safeguards is challenging, and many papers do not require this, but we encourage authors to take this into account and make a best faith effort.
    \end{itemize}

\item {\bf Licenses for existing assets}
    \item[] Question: Are the creators or original owners of assets (e.g., code, data, models), used in the paper, properly credited and are the license and terms of use explicitly mentioned and properly respected?
    \item[] Answer: \answerYes{} 
    \item[] Justification: All assets are properly credited, and licenses are mentioned.
    \item[] Guidelines:
    \begin{itemize}
        \item The answer NA means that the paper does not use existing assets.
        \item The authors should cite the original paper that produced the code package or dataset.
        \item The authors should state which version of the asset is used and, if possible, include a URL.
        \item The name of the license (e.g., CC-BY 4.0) should be included for each asset.
        \item For scraped data from a particular source (e.g., website), the copyright and terms of service of that source should be provided.
        \item If assets are released, the license, copyright information, and terms of use in the package should be provided. For popular datasets, \url{paperswithcode.com/datasets} has curated licenses for some datasets. Their licensing guide can help determine the license of a dataset.
        \item For existing datasets that are re-packaged, both the original license and the license of the derived asset (if it has changed) should be provided.
        \item If this information is not available online, the authors are encouraged to reach out to the asset's creators.
    \end{itemize}

\item {\bf New assets}
    \item[] Question: Are new assets introduced in the paper well documented and is the documentation provided alongside the assets?
    \item[] Answer: \answerNA{} 
    \item[] Justification: No new assets are released.
    \item[] Guidelines:
    \begin{itemize}
        \item The answer NA means that the paper does not release new assets.
        \item Researchers should communicate the details of the dataset/code/model as part of their submissions via structured templates. This includes details about training, license, limitations, etc. 
        \item The paper should discuss whether and how consent was obtained from people whose asset is used.
        \item At submission time, remember to anonymize your assets (if applicable). You can either create an anonymized URL or include an anonymized zip file.
    \end{itemize}

\item {\bf Crowdsourcing and research with human subjects}
    \item[] Question: For crowdsourcing experiments and research with human subjects, does the paper include the full text of instructions given to participants and screenshots, if applicable, as well as details about compensation (if any)? 
    \item[] Answer: \answerNA{} 
    \item[] Justification: The paper does not involve crowdsourcing nor research with human subjects.
    \item[] Guidelines:
    \begin{itemize}
        \item The answer NA means that the paper does not involve crowdsourcing nor research with human subjects.
        \item Including this information in the supplemental material is fine, but if the main contribution of the paper involves human subjects, then as much detail as possible should be included in the main paper. 
        \item According to the NeurIPS Code of Ethics, workers involved in data collection, curation, or other labor should be paid at least the minimum wage in the country of the data collector. 
    \end{itemize}

\item {\bf Institutional review board (IRB) approvals or equivalent for research with human subjects}
    \item[] Question: Does the paper describe potential risks incurred by study participants, whether such risks were disclosed to the subjects, and whether Institutional Review Board (IRB) approvals (or an equivalent approval/review based on the requirements of your country or institution) were obtained?
    \item[] Answer: \answerNA{} 
    \item[] Justification: The paper does not involve crowdsourcing nor research with human subjects.
    \item[] Guidelines:
    \begin{itemize}
        \item The answer NA means that the paper does not involve crowdsourcing nor research with human subjects.
        \item Depending on the country in which research is conducted, IRB approval (or equivalent) may be required for any human subjects research. If you obtained IRB approval, you should clearly state this in the paper. 
        \item We recognize that the procedures for this may vary significantly between institutions and locations, and we expect authors to adhere to the NeurIPS Code of Ethics and the guidelines for their institution. 
        \item For initial submissions, do not include any information that would break anonymity (if applicable), such as the institution conducting the review.
    \end{itemize}

\item {\bf Declaration of LLM usage}
    \item[] Question: Does the paper describe the usage of LLMs if it is an important, original, or non-standard component of the core methods in this research? Note that if the LLM is used only for writing, editing, or formatting purposes and does not impact the core methodology, scientific rigorousness, or originality of the research, declaration is not required.
    \item[] Answer: \answerYes{} 
    \item[] Justification: All usage of LLMs (in computing similarities and in pretraining) are clearly described in the paper.
    \item[] Guidelines:
    \begin{itemize}
        \item The answer NA means that the core method development in this research does not involve LLMs as any important, original, or non-standard components.
        \item Please refer to our LLM policy (\url{https://neurips.cc/Conferences/2025/LLM}) for what should or should not be described.
    \end{itemize}

\end{enumerate}

\end{document}